\documentclass{article}

\usepackage{arxiv}

\usepackage[utf8]{inputenc} 
\usepackage[T1]{fontenc}    
\usepackage{hyperref}       
\usepackage{url}            
\usepackage{booktabs}       
\usepackage{amsfonts}       
\usepackage{nicefrac}       
\usepackage{microtype}      
\usepackage{lipsum}		
\usepackage{graphicx}
\usepackage{natbib}
\usepackage{doi}

\usepackage{etoolbox}
\usepackage{array}
\usepackage{stfloats}
\usepackage{hyperref}
\usepackage{xurl}
\usepackage{verbatim}
\usepackage{amsthm}
\usepackage{amssymb}
\usepackage{dsfont}
\usepackage{multirow}
\usepackage{bm}
\usepackage{mathtools}
\usepackage{xspace}
\usepackage[section]{algorithm}
\usepackage[noend]{algpseudocode}
\usepackage{relsize}
\usepackage{xr}
\usepackage {diagbox}
\usepackage{booktabs,arydshln}
\usepackage{colortbl}
\usepackage{hhline}
\usepackage{caption}
\usepackage{subcaption}
\usepackage{mathtools}
\usepackage{adjustbox}
\usepackage{makecell}
\mathtoolsset{showonlyrefs=true}

\makeatletter
\DeclareRobustCommand\onedot{\futurelet\@let@token\@onedot}
\def\@onedot{\ifx\@let@token.\else.\null\fi\xspace}
\def\eg{e.g\onedot} 
\def\ie{i.e\onedot}

\def\etal{et al\onedot} 
\makeatother

\newcommand{\Tref}[1]{Table~\ref{#1}}
\newcommand{\Eref}[1]{Equation~\eqref{#1}}
\newcommand{\Fref}[1]{Figure~\ref{#1}}
\newcommand{\Sref}[1]{Section~\ref{#1}}
\newcommand{\Appendixref}[1]{Appendix~\ref{#1}}
\newcommand{\Aref}[1]{Algorithm~\ref{#1}}
\newcommand{\Lref}[1]{line~\ref{#1}}
\newcommand{\argmax}{\operatornamewithlimits{argmax}}

\algrenewcommand\algorithmicindent{0.7em}%

\title{Distilling Monolingual and Crosslingual Word-in-Context Representations}

\author{ \href{https://yukiar.github.io/}{Yuki Arase}
\thanks{This study extended the authors' paper \cite{arase:emnlp-findings2021} to conduct crosslingual word-in-context representation distillation.} \\
	School of Computing\\
	Tokyo Institute of Technology, Japan\\
	\texttt{arase@c.titech.ac.jp} \\
	\And
	\href{https://scholar.google.com/citations?user=cCAR9aYAAAAJ}{Tomoyuki Kajiwara} \\
	Graduate School of Science and Engineering\\ 
        Ehime University, Japan\\
	\texttt{kajiwara@cs.ehime-u.ac.jp} \\
}

\date{}

\renewcommand{\shorttitle}{Distilling Monolingual and Crosslingual Word-in-Context Representations}

\hypersetup{
pdftitle={Distilling Monolingual and Crosslingual Word-in-Context Representations},
pdfsubject={Distilling Monolingual and Crosslingual Word-in-Context Representations},
pdfauthor={Yuki Arase, Tomoyuki Kajiwara},
pdfkeywords={Word embedding, Word-in-Context Representation, Lexical semantics},
}

\begin{document}
\maketitle

\begin{abstract}
In this study, we propose a method that distils representations of word meaning in context from a pre-trained masked language model in both monolingual and crosslingual settings. 
Word representations are the basis for context-aware lexical semantics and unsupervised semantic textual similarity (STS) estimation. 
Different from existing approaches, our method does not require human-annotated corpora nor updates of the parameters of the pre-trained model. 
The latter feature is appealing for practical scenarios where the off-the-shelf pre-trained model is a common asset among different applications.  
Specifically, our method learns to combine the outputs of different hidden layers of the pre-trained model using self-attention. 
Our auto-encoder based training only requires an automatically generated corpus. 
To evaluate the performance of the proposed approach, we performed extensive experiments using various benchmark tasks. 
The results on the monolingual tasks confirmed that our representations exhibited a competitive performance compared to that of the previous study for the context-aware lexical semantic tasks and outperformed it for STS estimation. 
The results of the crosslingual tasks revealed that the proposed method largely improved crosslingual word representations of multilingual pre-trained models. 
\end{abstract}


\section{Introduction}
Word representations are the basis for various natural language processing (NLP) tasks. 
Particularly, they are crucial components in context-aware lexical semantics and in the estimation of unsupervised semantic textual similarity (STS) in both monolingual~\cite{ethayarajh-2018-unsupervised,yokoi-etal-2020-word,liu-etal-2020-towards-better} and crosslingual~\cite{wieting-etal-2020-bilingual,msbert} settings. 
Word representations are expected to represent the word meaning in context to improve these downstream tasks. 
Large-scale masked language models pre-trained on massive corpora, \eg, bi-directional encoder representations from transformers (BERT)~\cite{bert}, embed both the context and meaning of a word. Therefore, word-level representations generated by such masked language models are known as contextualised word representations. 
Previous studies~\cite{ethayarajh-2019-contextual,vulic-etal-2020-probing} have revealed that lexical and context-specific information is captured in different layers of masked language models. 
They argue that a sophisticated mechanism is required to derive the representations of word meaning in context from them. 
Although contextualised word representations have shown considerable promise, how best to compose the outputs of different layers of masked language models to effectively represent the word meaning in context remains an open question.

Similar to other NLP tasks, fine-tuning the pre-trained model with human-annotated corpora is a promising approach \cite{superglue}. 
However, annotating word meanings in context is non-trivial, and no such resource is abundantly available. 
Existing studies \cite{mirrorbert,mirrorwic} applied contrastive learning \cite{gao-etal-2021-simcse} to fine-tune the pre-trained masked language model without human-annotated corpora, which tweaks the model to generate representations for word meaning in context. 
One drawback of this approach is that the fine-tuned model is no longer easy to apply to other tasks, which is widely known as the catastrophic forgetting phenomenon \cite{Kirkpatrick2016OvercomingCF,lee-et-al-2017-catastrophic-forgetting}. 
Considering that the pre-trained model has become the basic building block of various NLP applications, keeping the model intact and reusing it for various tasks is appealing in practical scenarios. 
As a representative study of this direction, Liu~\etal~\cite{liu-etal-2020-towards-better} learn the transformation of the space of contextualised word representations towards static word embedding, \eg, fastText~\cite{fasttext} without updating the parameters of the pre-trained model. 
An adverse effect of this approach is that the transformation may excessively weaken the context information useful for composing the representations.

To address these challenges, we have proposed a method that distils representations of word meaning in context from the off-the-shelf masked language model \cite{arase:emnlp-findings2021}.\footnote{The code is available at \url{https://github.com/yukiar/distil_wic}.} 
We designed the distillation model as an autoencoder that reconstructs original representations, which allows training without human-annotated corpora.
Specifically, it learns to combine the outputs of different hidden layers using a self-attention mechanism~\cite{attention_is_all_you_need}. 
This study has extended our previous method for distilling crosslingual word-in-context representations. 
Contrary to the transformation-based approach, our representations preserve useful context information for composing word meaning in context.

We conducted evaluations on various benchmark tasks in monolingual and crosslingual settings. 
The monolingual results showed that our representations exhibited a performance competitive with that of the state-of-the-art method that transformed contextualised representations for context-aware lexical semantics. 
Moreover, the results confirm that our representations are more effective for composing sentence representations, which contributes to unsupervised STS estimation. 
Furthermore, the crosslingual evaluation results reveal that our method successfully distils representations of word meaning in context across languages, which enables efficient crosslingual similarity estimation.

\section{Related Work}
This section reviews studies on representation learning for word meaning in context. 
It also discusses the disentanglement of representations that is closely related to the present study methodologically. 

\subsection{Representation Learning for Word Meaning in Context}
\label{sec:related_work_wic_embedding}
An intuitive approach for obtaining representations of word-in-context might be to supervise a pre-trained masked language model using human-annotated corpora. 
While effective \cite{superglue,mcl-wic}, the annotation of word meaning in context is non-trivial, which hinders the abundant production of such corpora. 
To address this problem, previous studies \cite{mirrorbert,mirrorwic} applied contrastive learning to fine-tune a pre-trained model without human-annotated corpora. 
However, the parameter update during training may cause catastrophic forgetting, which makes reusing the fine-tuned model for other NLP tasks difficult. 

An orthogonal approach trains an independent model that \emph{transforms} representations generated by the frozen pre-trained model. 
This approach allows reusing the off-the-shelf pre-trained model as a common asset among different NLP applications. 
The transformation has been used to adjust the excessive effects of the context that dominates the representations. 
Shi~\etal~\cite{shi-etal-2019-retrofitting} added a transformation matrix on top of the embedding layer of ELMo~\cite{elmo}. 
Their approach derives the matrix that makes final representations of the same words in paraphrased sentences similar, whereas those of non-paraphrases become distant. 
The method most relevant to the present study was proposed by Liu~\etal~\cite{liu-etal-2020-towards-better}.  
They transformed the space of word representations towards the rotated space of static word embedding using a crosslingual alignment technique~\cite{doval-etal-2018-improving} for context-aware lexical semantic tasks. 
In principle, these previous studies aim to make contextualised representations less sensitive to contexts through transformation and prevent them from dominating the representations. 
In contrast, we derive word-in-context representations by combining different layers of a pre-trained model while preserving useful context information. 
Furthermore, our method is applicable to obtain crosslingual word-in-context representations with minimum modifications, whereas these previous studies are not.

\subsection{Representation Disentanglement}
\label{sec:related_work_disentangle}
Disentanglement techniques are also relevant to our approach. 
They generate specialised representations dedicated to a specific aspect. 
Previous studies typically employed autoencoders, with the encoder and decoder learning to disentangle and reconstruct original representations, respectively. 
Wieting~\etal~\cite{wieting-etal-2020-bilingual} and Tiyajamorn~\etal~\cite{nattapoing:emnlp2021} disentangled language-specific styles and sentence meanings. 
Shen~\etal~\cite{shen_nips_2017} disentangled content and sentiment, whereas John~\etal~\cite{john-etal-2019-disentangled} and Cheng~\etal~\cite{cheng-etal-2020-improving} disentangled content and style. 
Moreover, Chen~\etal~\cite{chen-etal-2019-multi} disentangled semantics and syntax. 
Similarly, the technique of information bottleneck was applied to pre-trained models to derive specialised representations for specific tasks~\cite{li-eisner-2019-specializing} although it requires an annotated corpora for training. 
The removal of specific attributes from representations is also relevant. 
Previous studies have proposed methods for removing predetermined attributes instead of disentangling for multi-linguality~\cite{chen-etal-2018-adversarial,lample2018word} and debiasing~\cite{pmlr-v28-zemel13,barrett-etal-2019-adversarial}.

These previous studies assume that disentangled attributes are distinctive. 
For instance, language-dependent styles and meanings are supposed to be independent of each other. 
Similarly, studies on attribute removal assume that the removed attributes are independent of the information remaining in the output representations. 
In contrast, the distillation of word meaning in context requires a subtle balance because the context information should be present in the meaning representations if it is crucial for determining the word meaning.

\section{Distilling Word Meaning in Context}
\label{sec:model}
\begin{figure}[t!]
\centering
\includegraphics[width=0.4\linewidth]{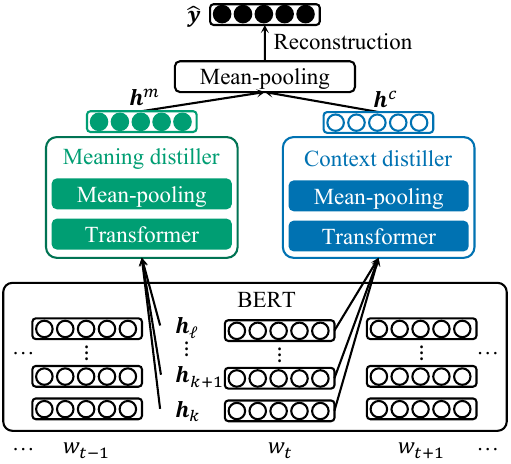}
\caption{Distillation of word meaning in context via an autoencoder}
\label{fig:ae}
\end{figure}

Inspired by the representation disentanglement approach discussed in \Sref{sec:related_work_disentangle}, we model the distillation of representations of word meaning in context using an autoencoder, as shown in \Fref{fig:ae}. 
This is a common architecture for both monolingual and crosslingual settings. 
Vuli{\'c}~\etal~\cite{vulic-etal-2020-probing} probed pre-trained language models for lexical semantic tasks, revealing that lexical information is scattered across lower layers whereas context-specific information is embedded in higher layers. 
Therefore, we aim to distil the outputs of different hidden layers using a transformer encoder layer. 

\Fref{fig:ae} shows the model architecture. 
First, we obtain the outputs of all the hidden layers of a masked language model, $\text{MLM}(\cdot)$, with frozen parameters $\bm{H}=\text{MLM}(S) \in \mathbb{R}^{|S| \times (\ell+1) \times d}$, where $S$ is an input sentence of length $|S|$ containing the target word; $w_t \in S$, $\ell$ is the number of hidden layers in the masked language model ($0$ corresponding to its embedding layer); $d$ is the hidden dimension of the masked language model. 
Thereafter, we extract the outputs of the hidden layers corresponding to the target word, $w_t$, from $\bm{H}$, indicating that $\bm{H}_{w_t}=[\bm{h}_0,\bm{h}_1,\cdots,\bm{h}_\ell]^\intercal \in \mathbb{R}^{(\ell+1) \times d}$. 
When $w_t$ is segmented into a set of $m$ sub-words $\tau_1, \tau_2, \cdots, \tau_m$, by a tokenizer of the masked language model, we compute the layer-wise averages of the hidden outputs of all the sub-words~\cite{bommasani-etal-2020-interpreting}. 
This indicates that $\bm{h}_i \in \bm{H}_{w_t}$ becomes 
\begin{equation}
    \bm{h}_i=\text{Pool}(\bm{h}_i^{\tau_1},\cdots, \bm{h}_i^{\tau_m}), 
\end{equation}
where $\bm{h}_i^{\tau_j}$ is the $i$th hidden output of a sub-word $\tau_j$ and the $\text{Pool}(\cdot)$ function conducts mean-pooling.

Thereafter, we input these hidden outputs into a meaning distillation model to derive a representation of word meaning in context. 
We also input the hidden outputs to another distillation model that derives information other than word meaning in context. 
Hereinafter, we refer to this information as the \emph{context} and the distillation model as the context distillation model.\footnote{The context here should be a mixture of different information, characterizing the target word and sentence, such as the meaning of the entire sentence, syntax, etc.} 
Each distillation model consists of a transformer encoder layer followed by a mean-pooling function to obtain the meaning and context representations, which is expressed as $\bm{h}^m \in \mathbb{R}^d$ and $\bm{h}^c\in \mathbb{R}^d$, respectively.
\begin{align} 
    \hat{\bm{h}}_k,\hat{\bm{h}}_{k+1},\cdots,\hat{\bm{h}}_\ell &=\text{TransformerEnc}(\bm{h}_k,\bm{h}_{k+1},\cdots,\bm{h}_\ell),\\
    \bm{h}^m &=\text{Pool}(\hat{\bm{h}}_k,\hat{\bm{h}}_{k+1},\cdots,\hat{\bm{h}}_\ell),
\end{align}
where $k \in [0, \ell]$ determines the bottom layer to be considered and $\text{TransformerEnc}(\cdot)$ represents the transformer encoder layer.  
Similarly, we distil the context representation. 
Averaging the outputs of the layers in the top half of the masked language models consistently performs well for context-aware lexical semantic tasks~\cite{vulic-etal-2020-probing,liu-etal-2020-towards-better}. 
Thus, we set $k=\ell/2+1$ to use the top-half layers for distillation.\footnote{We also attempted to make $k=1$ for using all hidden layers, which demonstrated slightly inferior performance to the top-half setting.} 

Finally, we reconstruct the original representation from $\bm{h}^m$ and $\bm{h}^c$. 
Although there are different approaches for reconstructions, such as using a neural network-based decoder, a sophisticated decoder may learn to fit itself to replicate the masked language model outputs. 
Hence, we adopt mean-pooling as the simplest mechanism for reconstruction.
\begin{equation}
    \hat{\bm{y}}=\text{Pool}(\bm{h}^m, \bm{h}^c).
\end{equation}
The reconstruction target $\bm{y}\in \mathbb{R}^d$ is the mean-pooled hidden layers of the original masked language model. 
\begin{equation}
    \bm{y}=\text{Pool}(\bm{h}_{k},\bm{h}_{k+1}, \cdots,\bm{h}_\ell).\label{eq:y}
\end{equation}
We minimise the reconstruction loss as:
\begin{equation} \label{eq:loss_recon}
    \mathcal{L}_r=\frac{1}{d}\|\bm{y}-\hat{\bm{y}}\|^2_2.
\end{equation}
For inference, we use $\bm{h}^m$ as a representation of word meaning in context.

John~\etal~\cite{john-etal-2019-disentangled} reported that a variational autoencoder~\cite{vae} outperformed the simpler autoencoder on representation disentanglement. 
However, this was not the case in this study, where the autoencoder consistently outperformed the variational version. 
We intend to further investigate auto-encoding architectures in future studies.  

\section{Learning Framework}
\begin{figure*}
     \centering
     \begin{subfigure}[b]{0.48\textwidth}
         \centering
        \includegraphics[width=\textwidth]{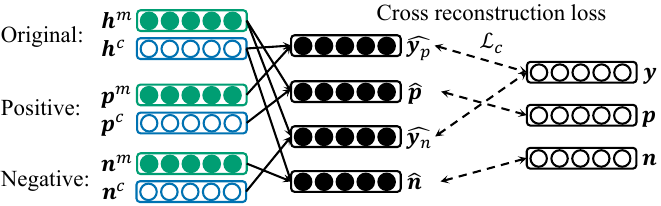}
         \caption{Monolingual}
         \label{fig:self_training:mono}
     \end{subfigure}
     \hfill
     \begin{subfigure}[b]{0.48\textwidth}
         \centering
         \includegraphics[width=\textwidth]{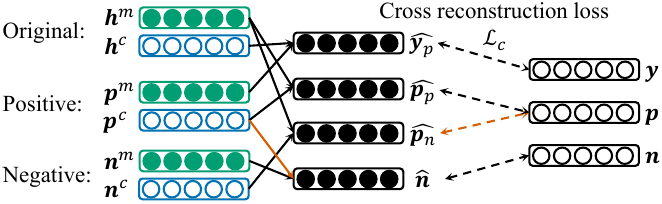}
         \caption{Crosslingual}
         \label{fig:self_training:xl}
     \end{subfigure}
        \caption{Cross reconstruction with automatically generated positive and negative samples (the orange arrows indicate differences between the monolingual and crosslingual settings).}
        \label{fig:self_training}
\end{figure*}

The meaning and context distillation models described in \Sref{sec:model} require constraints to ensure that the desired attributes are distilled; otherwise, these distillation models obtain a degenerated solution that simply copies the original representations. 
We design a framework ensuring that word meaning in context is distilled using an automatically generated training corpus. 

Subsequently, we first describe our learning framework for the monolingual settings (\Sref{sec:cross-reconstruction} and \Sref{sec:corpus}). 
Thereafter, we adapt the framework for the crosslingual settings (\Sref{sec:self-supervsion-xl}) with minimal modifications. 

\subsection{Cross Reconstruction}
\label{sec:cross-reconstruction}

\begin{table}[t!]
\centering
\caption{Training examples (\emph{Italicized} words represent $w_t$, $w_p$, and $w_n$, respectively.)}
\label{tab:example_input}
\begin{tabular}{@{}lp{0.85\linewidth}@{}}
\toprule
original      & Proceedings are all \emph{instituted} in the name of the queen [...]  \\
positive & All proceedings are \emph{initiated} on behalf of the Queen [...]     \\
negative & Proceedings are all \emph{made} in the name of the queen [...] \\ \midrule
original      & The by-election was held due to the \emph{incumbent} Irish Parliamentary MP , [Person Name] , being declared bankrupt .   \\
positive &  The secondary elections were held because the \emph{current} Member of the Irish Parliament , [Person Name] , was declared bankrupt .    \\
negative &  The by-election was held due to the \emph{previous} Irish Parliamentary MP , [Person Name] , being declared bankrupt . \\ \midrule
original      & [...] they have paid for the \emph{requisite} amount of tonnes to be recovered each year .  \\
positive & [...] they have paid for the \emph{necessary} quantity of tonnes , which must be recovered every year .    \\
negative & [...] they have paid for the \emph{proportioned} amount of tonnes to be recovered each year . \\
\bottomrule
\end{tabular}%
\end{table}

Assume we have two sentences, $S_p$ and $S_n$, as positive and negative examples, respectively, as shown in \Tref{tab:example_input}.
$S_p$ is a paraphrase of $S$ that must contain a word $w_p$, which is equivalent to $w_t$ or a lexical paraphrase of $w_t$. 
In contrast, $S_n$ is a negative example that replaces $w_t$ with a word $w_n$ in $S$, which has a different meaning from $w_t$ but fits in the context of $S$. 

Using the hidden outputs of $w_p$ and $w_n$, we distil the meaning and context representations, $\bm{p}^m$ and $\bm{p}^c$, and those of $\bm{n}^m$ and $\bm{n}^c$, respectively. 
The meaning representation of $w_t$, $\bm{h}^m$, should satisfy the following two conditions. 
\begin{itemize}
    \item $\bm{h}^m$ can be combined with $\bm{p}^c$ to reconstruct the original representation derived for $w_p$. 
    \item $\bm{h}^m$ can be combined with $\bm{n}^c$ to reconstruct the original representation, $\bm{y}$. 
\end{itemize}
Similarly, the context representation, $\bm{h}^c$, should satisfy the following two conditions.
\begin{itemize}
    \item $\bm{h}^c$ can be combined with $\bm{p}^m$ to reconstruct the original representation, $\bm{y}$.
    \item $\bm{h}^c$ can be combined with $\bm{n}^m$ to reconstruct the original representation derived for $w_n$. 
\end{itemize}
We use these properties of meaning and context representations as constraints. 

Specifically, we train the model to achieve the cross-reconstruction of meaning and context representations as depicted in \Fref{fig:self_training:mono}. 
\begin{align}
    \hat{\bm{p}} &=\text{Pool}(\bm{h}^m, \bm{p}^c),
    & \hat{\bm{y}_p}=\text{Pool}(\bm{p}^m, \bm{h}^c),\\
    \hat{\bm{n}} &=\text{Pool}(\bm{n}^m, \bm{h}^c),
    & \hat{\bm{y}_n}=\text{Pool}(\bm{h}^m, \bm{n}^c). \label{eq:loss_cross_terms}
\end{align}
Our objective minimises the following cross-reconstruction loss as expressed below:
\begin{equation} \label{eq:loss_cross}
        \mathcal{L}_c=\frac{1}{d}\{ \|\bm{p}-\hat{\bm{p}}\|^2_2  + \|\bm{y}-\hat{\bm{y}_p}\|^2_2 +\|\bm{n}-\hat{\bm{n}}\|^2_2+\|\bm{y}-\hat{\bm{y}_n}\|^2_2\},
\end{equation}
where $\bm{p}$ and $\bm{n}$ are computed similarly to \Eref{eq:y}. 
The overall loss function is the summation of the reconstruction and cross-reconstruction losses in Equations~\eqref{eq:loss_recon} and \eqref{eq:loss_cross}
\begin{equation}
    \mathcal{L}=\mathcal{L}_r+\mathcal{L}_c,
\end{equation}
where $\mathcal{L}_r$ is expanded to sum the reconstruction losses of the positive and negative samples. 

\subsection{Training Corpus Creation}
\label{sec:corpus}
In this section, we describe the generation of a training corpus in an automatic way using techniques of round-trip translation, word alignment, and masked token prediction. 

\paragraph{Round-trip Translation and Word Alignment}
As a positive example $S_p$, we need a paraphrase of $S$ containing a word $w_p$ that is a lexical paraphrase of $w_t$. 
First, we automatically generate paraphrases using round-trip translation \cite{Kajiwara_Miura_Arase_2020}. 
Next, we conduct word alignment to identify $w_p$ in a paraphrased sentence. 
Specifically, we adopt a simple word alignment method commonly used in previous studies \cite{jalili-sabet-etal-2020-simalign,garg-etal-2019-jointly,och-ney-2003}, which aligns a word $w_i$ to $w_j$ if and only if $w_i$ is most similar to $w_j$ and vice-versa. 
The similarity is computed by cosine similarities between their embeddings.  
We add heuristics to the word alignment method to improve alignment accuracy as detailed in \Appendixref{appendix:word_alignment}. 
Note that candidates in which $w_p$ is not identified are discarded.

\paragraph{Masked Token Prediction}
Negative samples replace $w_t$ with an arbitrary word $w_n$ that fits in the context of $S$. 
We generate candidates for the replacement of words using the masked token prediction, which is the primary task used to train the masked language model. 
Specifically, we input an original sentence whose target is masked by the {\tt [MASK]} label to the masked language model, and we obtain predictions $T=\{t_1, \cdots, t_{|V|}\}$ with probabilities, $Q=\{q_1,\cdots, q_{|V|}\}$, where $|V|$ is the size of the vocabulary of the masked language model. 
To avoid selecting a possible paraphrase of $w_t$ as $w_n$, we refer to the static word-embedding model following Qiang~\etal~\cite{BERT_LS}. 
We sort $T$ in descending order of $Q$ and identify $w_n$ with the lower cosine similarity to $w_t$ based on heuristics detailed in \Appendixref{appendix:masked_token_prediction}.

We apply the same technique to enhance $w_p$ when it is identical or similar to $w_t$ based on a character-level edit distance. 
Where possible, we replace $w_p$ with $w'_{p} \in T$ when it has high cosine similarity to $w_t$.

\subsection{Training for Crosslingual Scenario}
\label{sec:self-supervsion-xl}

\begin{table}[t!]
\centering
\caption{Training examples for crosslingual settings (\emph{Italicized} words represent $w_t$, $w_p$, and $w_n$, respectively.)}
\label{tab:example_input_xl}
\begin{tabular}{@{}clp{0.75\linewidth}@{}}
\toprule
      & original & Meltdown could potentially impact [...] computers than \emph{presently} identified [...]            \\
en-es & positive & Potencialmente Meltdown puede afectar [...] dispositivos mayor que la identificada \emph{actualmente} [...]   \\
      & negative & Potencialmente Meltdown puede afectar [...] dispositivos mayor que la identificada \emph{inicialmente} [...]           \\ \midrule
      & original & In $1999$, he left the project to \emph{pursue} a solo career. \\
en-de & positive & Im Jahr $1999$ verlie\ss er das Projekt, um eine Solokarriere zu \emph{verfolgen} . \\
      & negative & Im Jahr $1999$ verlie\ss er das Projekt, um eine Solokarriere zu \emph{beginnen} .  \\ \midrule
      & original & [Person Name] boasted to his \emph{government} that [...]       \\
en-et & positive & [Person Name] uhkustas oma \emph{valitsuse} ees, [...]         \\
      & negative & [Person Name] uhkustas oma \emph{tellijate} ees, [...]           \\ 
\bottomrule
\end{tabular}%
\end{table}

Because mutual translations are paraphrases in the crosslingual case, we can employ off-the-shelf parallel corpora. 
We adapt the learning framework for the crosslingual settings such that the differences in languages, \eg, language-specific styles, are also distilled into the context representations. 
For this purpose, we regard English sentences as original and use their parallel (non-English) sentences both for positive and negative samples. 
Specifically, we identify lexical translations of the target words on parallel sentences as positive samples. 
Besides, we generate negative samples based on the positive samples using masked token prediction, in contrast to the monolingual settings wherein we generated negative samples based on the original sentences. 
\Tref{tab:example_input_xl} shows examples of our crosslingual training corpus.

As illustrated in \Fref{fig:self_training:xl}, where orange arrows indicate changes from the monolingual version, we reconstruct the original, positive, and negative representations as follows:
\begin{align}
    \hat{\bm{p}_p} &=\text{Pool}(\bm{h}^m, \bm{p}^c),
    & \hat{\bm{y}_p}=\text{Pool}(\bm{p}^m, \bm{h}^c),\\
    \hat{\bm{p}_n} &=\text{Pool}(\bm{h}^m, \bm{n}^c),
    & \hat{\bm{n}}=\text{Pool}(\bm{n}^m, \bm{p}^c). 
\end{align}
The first two reconstructions are common with the monolingual case (\Eref{eq:loss_cross_terms}). 
Because of the language difference, the meaning representation of $w_t$, $\bm{h}^m$, and the context representation of $w_n$, $\bm{n}^c$, should reconstruct the original representation of $w_p$ in the same language with $w_n$. 
Similarly, the meaning representation of $w_n$, $\bm{n}^m$, and the context representation of the $w_p$, $\bm{p}^c$, should reconstruct the original representation of $w_n$. 
Finally, the cross-reconstruction loss (\Eref{eq:loss_cross}) becomes
\begin{equation} 
        \mathcal{L}_c=\frac{1}{d}\{\|\bm{p}-\hat{\bm{p}_p}\|^2_2  + \|\bm{y}-\hat{\bm{y}_p}\|^2_2+\|\bm{p}-\hat{\bm{p}_n}\|^2_2 +\|\bm{n}-\hat{\bm{n}}\|^2_2\}.
\end{equation}

We identified positive and negative correspondence to $w_t$, $w_p$ and $w_n$, respectively, for the crosslingual settings using the same procedure with the monolingual case on the parallel corpora as detailed in \Appendixref{appendix:word_alignment} and \Appendixref{appendix:masked_token_prediction}. 
The conditions of $w_p$ and $w_n$ must satisfy are set stricter than those of the monolingual case because the quality of crosslingual static word embedding is less reliable compared to the monolingual counterpart. 

\section{Experimental Setup}
\label{sec:main_eval}
We empirically evaluated whether our method distils representations of word meaning in context from a masked language model using context-aware lexical semantic tasks and STS estimation tasks.\footnote{We list the URLs of all the dependent language resources, toolkits, and libraries in \Appendixref{appendix:resources}.} 
All the experiments were conducted on an NVIDIA Tesla V$100$ GPU. 

\subsection{Context-aware Lexical Semantic Tasks}
We followed the experimental settings used by Liu~\etal~\cite{liu-etal-2020-towards-better} for a fair and systematic performance comparison. 
They categorised monolingual context-aware lexical semantic tasks into {\bf Within-word} and {\bf Inter-word} tasks. 
The former evaluates the diversity of word representations for different meanings of the same word associated with different contexts. 
In contrast, the latter evaluates the similarity of word representations for different words when they have the same meaning. 
Besides, we evaluate the {\bf Crosslingual} tasks. 
The left-side columns of \Tref{tab:mono_corpora_stats} show the number of word pairs in the monolingual evaluation corpora and those of \Tref{tab:xl_corpora_stats} show the number of word pairs in the crosslingual evaluation corpus per language pair. 

\begin{table}[t!]
\centering
\caption{Statistics of the monolingual evaluation corpora}
\label{tab:mono_corpora_stats}
\begin{tabular}{@{}lclc@{}}
\toprule
Lexical Semantics & \# of pairs    & STS      & \# of pairs \\ \midrule
USim              & $1.1$k                 & STS 2012 & $3.1$k                 \\
WiC               & $1.4$k                 & STS 2013 & $1.5$k                 \\
CoSimlex-I        & \multirow{2}{*}{$680$} & STS 2014 & $3.7$k                 \\
CoSimlex-II       &                      & STS 2015 & $8.5$k                 \\
SCWS              & $2$k                   & STS 2016  & $9.2$k                 \\ \bottomrule
\end{tabular}%
\end{table}

\begin{table}[t!]
\centering
\caption{Statistics of the crosslingual evaluation corpora}
\label{tab:xl_corpora_stats}
\begin{tabular}{@{}lclc@{}}
\toprule
MCL-WiC & \# of pairs & STS $2017$ & \# of pairs \\ \midrule
en-ar  & $992$      & en-ar & $250$      \\
en-fr  & $990$      & en-de & $250$      \\
en-ru  & $992$      & en-tr & $250$      \\
en-zh  & $982$      & en-es & $250$      \\ \bottomrule
\end{tabular}%
\end{table}

\paragraph{Within-word Tasks}
The within-word evaluation was divided into three tasks. 
The first is based on the Usage Similarity ({\bf Usim}) corpus~\cite{erk-etal-2013-measuring}, which provides graded similarity between the meanings of the same word in a pair of different contexts. 
The second task uses the Word in Context ({\bf WiC}) corpus~\cite{pilehvar-camacho-collados-2019-wic}, which provides binary judgements to verify whether the meaning of a given word varies in different contexts. 
Following the standard setting recommended in the original study, we tuned the threshold for cosine similarity between word representations to make binary judgments. 
Specifically, we searched the threshold in the range of $[0, 1.0]$ with $0.01$ intervals to maximise the accuracy of the validation set. 
The third task is the subtask-$1$ of CoSimlex~\cite{armendariz-etal-2020-cosimlex} (denoted as {\bf CoSimlex-I}). 
CoSimlex provides a pair of contexts consisting of several sentences for each word pair extracted from SimLex-$999$~\cite{hill-etal-2015-simlex}. 
It annotates the graded similarity in each context. 
CoSimlex-I requires the estimation of the \emph{change} in similarities between the same word pair in different contexts. 
Hence, it evaluates whether representations can change for different word meanings based on the context. 

\paragraph{Inter-word Tasks}
The inter-word evaluation consisted of two tasks. 
The first was the subtask-$2$ of CoSimlex (denoted as {\bf CoSimlex-II}), which required estimating the similarity between different word pairs in the same context. 
The second task used the Stanford Contextual Word Similarity ({\bf SCWS}) corpus~\cite{huang-etal-2012-improving}, which provided graded similarity between word pairs in a pair of different contexts. 
The contexts of CoSimlex and SCWS consist of several sentences. 
We input all the sentences as a single context. 

\paragraph{Crosslingual WiC Tasks}
We used the Multilingual and Crosslingual Word-in-Context ({\bf MCL-WiC}) corpus~\cite{mcl-wic}, which provides test sets for English-Arbic (en-ar), English-French (en-fr), English-Russian (en-ru), and English-Chinese (en-zh) language pairs.  
The MCL-WiC task requires to determine whether an English word in a sentence and its lexical translation in a sentence of other language share the same meaning. 
Although the original MCL-WiC corpus allowed an English word to be translated into a multi-word expression and periphrasis, we excluded pairs in which a target splits into non-consecutive phrases in the other language. 
We similarly made binary judgements with the monolingual WiC task. 
The original corpus covers both multilingual and crosslingual tasks, and provides validation sets only for the multilingual tasks. 
Hence, following the organisers' report~\cite{mcl-wic}, we used the corresponding multilingual validation sets in languages other than English to determine the threshold for binarization. For instance, we used the `fr-fr' validation set for the `en-fr' task. 

\paragraph{Evaluation Metrics}
We estimated the similarity between words using cosine similarity in their representations. 
We used the official evaluation metrics determined by each corpus. 
Namely, we evaluated WiC and MCL-WiC using accuracy, CoSimlex-I using Pearson's $r$, and others using Spearman's $\rho$. 
Note that the performance on the WiC test set was measured on the CodaLab server.\footnote{\url{https://competitions.codalab.org/competitions/20010}}

\subsection{STS Tasks}
We also evaluated the proposed method on the STS tasks. 
Cosine similarity is commonly used to estimate the similarity between two text representations. 
We also used cosine similarity because such a primitive measure is sensible to characteristics of different representations. 
We generated a sentence representation by simply averaging the representations of sub-words in a sentence, excluding the representations for special tokens preserved in BERT, \ie, {\tt [CLS]}, {\tt [SEP]}, and {\tt [PAD]}. 
Thereafter, we computed the cosine similarities between them.  

Regarding the monolingual tasks, we evaluated the $2012$ to $2016$ SemEval STS shared tasks~\cite{agirre-etal-2012-semeval,agirre-etal-2013-sem,agirre-etal-2014-semeval,agirre-etal-2015-semeval,agirre-etal-2016-semeval} ({\bf STS $\bf{2012}$ to $\bf{2016}$}) to predict the human scores that indicate the degree of semantic similarity between the two sentences. 
We downloaded and pre-processed the datasets using the SentEval toolkit~\cite{conneau-kiela-2018-senteval}. 
The right-side columns of \Tref{tab:mono_corpora_stats} show the number of sentence pairs in these corpora.

Considering a crosslingual task, we evaluated the extended version of SemEval $2017$ crosslingual STS shared task \cite{cer-etal-2017-semeval} ({\bf STS} $\bf{2017}$).  
Although the task consists of $7$ subtasks of different language pairs, we chose subsets whose sentences are created by human translations to exclude biases originating from specific machine translators: English-Arabic (en-ar), English-German (en-de), English-Turkish (en-tr), and English-Spanish (en-es).\footnote{Other three language pairs were created by machine translation.}
The right-side columns of \Tref{tab:xl_corpora_stats} show the number of sentence pairs in the STS~$2017$ corpus per language pair.

The Pearson's $r$ between the model predictions and human scores was used as an evaluation metric. 
Each STS corpus is divided by data sources. 
Hence, the corpus level score is the average of the Pearson's $r$ for each sub-corpus.


\subsection{Training Corpus Preparation}
Wikipedia is a common data source used to create training corpora for monolingual and crosslingual settings. 
Regarding both settings, we randomly sampled and excluded approximately $1\%$ sentences as the validation set and used the rest for the training.

\paragraph{Monolingual Training Corpus}
To prepare the training corpus for the monolingual settings, we used English Wikipedia dumps distributed for the WMT$20$ competition. 
The texts were extracted using WikiExtractor. 
As a pre-processing step, we first identified the language of each text using the langdetect toolkit and discarded all the non-English texts. 
Thereafter, we conducted sentence segmentation and tokenization using Stanza~\cite{qi-etal-2020-stanza} and extracted sentences of $15$ to $50$ words.

Considering the candidate target words, we extracted the top-$50$k frequent words\footnote{We excluded the top $0.1\%$ vocabulary because most were function words.} following Liu~\etal~\cite{liu-etal-2020-towards-better}. 
Thereafter, we sampled $1$M sentences containing these words from the pre-processed Wikipedia corpus. 
Using these $1$M sentences, we obtained paraphrases via round-trip translation where the translators were trained with exactly the same settings as Kajiwara~\etal~\cite{Kajiwara_Miura_Arase_2020}. 
Subsequently, we generated positive and negative samples using word alignment and masked token prediction, where we used fastText and BERT-Large, cased model, respectively. 

Round-trip translation does not always produce an alignable $w_p$, and our simple word alignment heuristic may fail to identify $w_p$.  
Hence, the final number of sentences in our training corpus was reduced to $929,265$, where $44,614$ unique words remained as targets. 
Among them, $242,643$ sentences contained $w_p$ whose surfaces were larger than the $3$ character-level edit distance, which were expected as lexical paraphrases. 
We used these $929$k triples of the original, positive, and negative samples.

\paragraph{Cross-Lingual Training Corpus}
\begin{table}[t!]
\centering
\caption{Number of triples in our crosslingual corpus per language pair}
\label{tb:crosslingual_corpus_stats}
\begin{tabular}{@{}cr@{}}
\toprule
Language pair & \multicolumn{1}{l}{Number of triples} \\ \midrule
en-de         & $706,308$                        \\
en-es         & $1,364,647$                      \\
en-fr         & $1,232,212$                      \\
en-ru         & $522,883$                        \\
en-ar         & $525,589$                        \\
en-ro         & $529,788$                        \\
en-tr         & $310,341$                        \\
en-et         & $165,585$                        \\
en-zh         & $166,726$                        \\ \midrule
Total         & $5,524,079$                      \\ \bottomrule
\end{tabular}%
\end{table}

To prepare a crosslingual training corpus, we used WikiMatrix~\cite{wikimatrix}, parallel corpora mined from the textual content of Wikipedia. 
Following Reimers and Gurevych~\cite{msbert}, we extracted parallel sentences, thresholding the margin score by $1.05$ to obtain the mutual translations. 
Thereafter, we applied the same pre-processing of language identification and tokenization as the monolingual settings, and obtained $50$k target words in English for each language pair. 

For conducting the word alignment and masked token prediction, we used aligned fastText~\cite{aligned-fasttext} and XLM-RoBERTa-Large model \cite{xlm-roberta}, respectively.
We found that the stricter heuristics in the word alignment and masked token prediction filtered out the majority of the candidates. 
Hence, we did not conduct sampling; however, we simply limited the number of sentences per target to at most $100$ to avoid frequent words from excessively dominating the training corpus. 

We constructed the training corpus to cover $9$ language pairs of various sizes and diverse language families. 
Consequently, our corpus consists of $5.5$M triples of the original, positive, and negative samples as shown in \Tref{tb:crosslingual_corpus_stats}. 
We used the English sentences as original and their parallel (non-English) sentences for positive and negative samples.

\begin{table*}[t!]
\centering
\caption{Results on the context-aware lexical-semantic tasks, where ``w/o NS'' denotes the proposed method without negative samples. The best scores are shown in \protect\textbf{bold} fonts and scores higher than BERT-Large are \underline{underlined} ($\rho$ represents Spearman's $\rho$ and $r$ denotes Pearson's $r$.).}
\label{tab:result_ls}
\begin{tabular}{@{}ccccccc@{}}
\toprule
\multicolumn{2}{l}{}                                  & \multicolumn{3}{c}{Within-Word} & \multicolumn{2}{c}{Inter-Word}         \\ \cmidrule(l){3-7} 
\multicolumn{2}{l}{}             & USim ($\rho$)  & WiC (accuracy ($\%$))   & CoSimlex-I ($r$) & CoSimLex-II ($\rho$) & SCWS ($\rho$)  \\ \midrule
\multicolumn{2}{l}{BERT-Large}   & $0.5966$ & $66.57$ & $0.7638$     & $0.7332$      & $0.7255$ \\
\multicolumn{2}{l}{Liu~\etal~\cite{liu-etal-2020-towards-better}}    & $\bm{0.6383}$    & $\bm{67.50}$    & $\bm{0.7710}$   & $0.7258$ & $\underline{0.7572}$                        \\ \midrule
                       & Meaning & $\underline{0.6305}$ & $\underline{67.29}$ & $0.7576$     & $\bm{0.7358}$      & $\bm{0.7594}$ \\
\multirow{-2}{*}{Proposed} & Context & $0.4147$ & $62.21$ & $0.6485$     & $0.5106$      & $0.2914$ \\
                       & Meaning & $0.2934$ & $57.79$ & $0.3843$     & $0.5022$      & $0.2883$ \\
\multirow{-2}{*}{w/o NS} & Context & $0.5929$    & $\underline{66.79}$    & $0.7617$   & $0.7296$ &  $\underline{0.7279}$ \\ \bottomrule
\end{tabular}%
\end{table*}

\begin{table}[t!]
\centering
\caption{Results of the STS $2012$ to $2016$ tasks, where ``w/o NS'' denotes our method without negative samples. The best scores are represented in \protect\textbf{bold} fonts, and the scores higher than BERT-Large are \underline{underlined}.}
\label{tab:result_sts}
\begin{tabular}{@{}cccccccc@{}}
\toprule
\multicolumn{2}{l}{} & STS12 & STS13 & STS14 & STS15 & STS16 & Average \\ \midrule
\multicolumn{2}{l}{BERT-large} & $0.572$ & $0.598$ & $0.629$ & $0.700$ & $0.675$ & $0.635$ \\
\multicolumn{2}{l}{Liu~\etal~\cite{liu-etal-2020-towards-better}} & $\underline{0.576}$ & $\underline{0.616}$ & $\underline{0.641}$ & $0.692$ & $\bm{0.687}$ & $\underline{0.642}$ \\ \midrule
\multirow{2}{*}{Proposed} & Meaning & $\bm{0.583}$ & $\bm{0.628}$ & $\bm{0.662}$ & $\bm{0.714}$ & $\underline{0.684}$ & $\bm{0.654}$ \\
 & Context & $0.460$ & $0.411$ & $0.466$ & $0.569$ & $0.575$ & $0.496$ \\
\multirow{2}{*}{w/o NS} & Meaning & $0.177$ & $0.181$ & $0.214$ & $0.238$ & $0.217$ & $0.205$ \\
 & Context & $\underline{0.573}$ & $\underline{0.602}$ & $\underline{0.635}$ & $\underline{0.706}$ & $\underline{0.683}$ & $\underline{0.640}$ \\ \bottomrule
\end{tabular}%
\end{table}

\subsection{Baselines}
Regarding the monolingual tasks, we compare our method to Liu~\etal~\cite{liu-etal-2020-towards-better} as the state-of-the-art in the family of methods that transform contextualised representations. 
Recall that Liu~\etal~\cite{liu-etal-2020-towards-better} transform representations from the masked language model using static word embeddings. 
Specifically, we used fastText as the static embeddings that performed most robustly across the models and tasks. 
We replicated their method using the implementation and training corpus published by the authors. 
Note that their training corpus was also drawn from English Wikipedia.

As discussed in \Sref{sec:related_work_wic_embedding}, there is no straightforward way to adapt Liu~\etal~\cite{liu-etal-2020-towards-better} to the crosslingual setting. 
Hence, we compared our method to Reimers and Gurevych~\cite{msbert}, one of the state-of-the-art methods for unsupervised crosslingual sentence representations (denoted as {\bf mSBERT} hereafter). 
They applied a knowledge distillation to construct a crosslingual model by forcing a representation of a translated sentence to be similar to the representation of the original sentence generated by a monolingual model.

We also compared our method to {\bf BERT-Large}, multilingual BERT ({\bf mBERT}), and XLM-RoBERTa ({\bf XLM-R}). 
Based on previous studies~\cite{vulic-etal-2020-probing,liu-etal-2020-towards-better}, we used the average of the outputs of the top-half layers on all the pre-trained models, \ie, \Eref{eq:y}, which consistently performed well in lexical semantic tasks.

\subsection{Implementation}
We implemented our method using PyTorch and Transformers library~\cite{wolf-etal-2020-transformers}. 
Regarding the monolingual tasks, we used the cased version of BERT-Large. 
For crosslingual tasks, we used `base' models for mBERT (the cased version), XLM-R, and mSBERT\footnote{Namely, we used {\tt paraphrase-xlm-r-multilingual-v1} that adopted XLM-RoBERTa-base for knowledge distillation.} owing to the larger training corpus size. 
Recall that the parameters of the pre-trained models were frozen and never fine-tuned. 

The meaning and context distillers of the proposed model included a transformer layer. 
We set the number of attention heads as eight.\footnote{We tried $16$ attention heads as in the BERT-Large model; however, the performance was comparable to that of $8$ heads.} 
We set the dimensions of the internal feed-forward network as four and six times of the input for the monolingual and crosslingual settings, respectively. 
We used the larger dimension for the latter because it had to deal with the extra-task to distil language-specific styles.  
We applied $10\%$ dropouts to the transformer layer. 
The batch sizes were $128$ and $512$ for the monolingual and crosslingual settings, respectively. 
We used AdamW~\cite{adamw} as an optimizer for which the learning rate was tuned on smaller samples \cite{learning_rate_finder}. 
For stable training, we applied a warm-up, where the initial learning rate was linearly increased for the first $1$k steps to reach the predetermined value. 
The training was stopped early with patience of $15$ and a minimum delta of $1.0e-5$ based on the validation loss. 

\section{Experimental Results}
Below, we discuss the experimental results of the monolingual and crosslingual tasks. 
We also analyse the results of the ablation study to investigate the contributions of the positive and negative samples. 

\subsection{Results on Monolingual Tasks}
\Tref{tab:result_ls} shows the results on the context-aware lexical semantic tasks. 
The superior performance of our meaning representation to context representations confirms that distillation has been performed as aimed.
Our meaning representations achieved a performance competitive with the transformation method by Liu~\etal~\cite{liu-etal-2020-towards-better}.\footnote{The performance of Liu~\etal~\cite{liu-etal-2020-towards-better} on CoSimlex-II and SCWS differed from their paper. The difference may be caused by the method used to compose a word-level representation when a word is segmented into sub-words. Because there is no explanation in their study, we have generated the word representation similarly, \ie, by layer-wise averaging of all the sub-words’ hidden outputs (also for the BERT baseline).} 
While the transformation method was stronger in the Within-Word tasks, our method outperformed it for Inter-Word tasks. 
This is because the transformation method makes representations of the same words in different contexts closer to the same static embedding; nonetheless, they do not explicitly model the relations across words. 
In contrast, our negative samples provide distant supervision, which makes the representations of words with different meanings distinctive. 
Although the performances of these two methods are competitive, these different properties are reflected in the representations. 

The difference is more pronounced in the results of unsupervised STS tasks shown in \Tref{tab:result_sts}.\footnote{The performance of the BERT-large baseline was improved from \cite{arase:emnlp-findings2021} by omitting paddings.} 
Our meaning representations outperformed the transformed representations in four out of the five tasks. 
The transformation has the effect of making contextualised representations less sensitive to contexts to prevent contexts from dominating the representations. 
This effect is preferred in tasks of context-aware lexical semantics that severely require representations of word meaning; meanwhile, it sacrifices context information valuable for STS. 
In contrast, our method does not waste the context information useful for composing sentence representations. 

\subsection{Results on Crosslingual Tasks}
\begin{table*}[t!]
\centering
\caption{Results on MCL-WiC tasks, where ``w/o NS'' denotes the proposed method without negative samples. The best test scores are shown in \protect\textbf{bold} fonts, and scores higher than the base pre-trained models are \underline{underlined}.}
\label{tab:mcl_wic_results}
\resizebox{\textwidth}{!}{%
\begin{tabular}{@{}ccccccccccccccc@{}}
\toprule
 &  & \multicolumn{3}{c}{en-ar} & \multicolumn{3}{c}{en-fr} & \multicolumn{3}{c}{en-ru} & \multicolumn{3}{c}{en-zh} & \multirow{2}{*}{Average} \\
 &  & Threshold & Dev & Test & Threshold & Dev & Test & Threshold & Dev & Test & Threshold & Dev & Test &  \\ \midrule
\multicolumn{2}{c}{Majority baseline} & \multicolumn{3}{c}{$50.10$} & \multicolumn{3}{c}{$50.00$} & \multicolumn{3}{c}{$50.10$} & \multicolumn{3}{c}{$50.31$} & $50.13$ \\ \midrule
\multicolumn{2}{c}{mBERT} & $0.782$ & $62.20$ & $50.15$ & $0.801$ & $69.80$ & $52.02$ & $0.776$ & $67.90$ & $51.41$ & $0.803$ & $65.40$ & $50.00$ & $50.90$ \\
Proposed & Meaning & $0.758$ & $62.20$ & $\underline{59.19}$ & $0.846$ & $68.70$ & $\underline{61.31}$ & $0.838$ & $67.70$ & $\underline{63.41}$ & $0.854$ & $63.00$ & $\underline{54.99}$ & $\underline{59.72}$ \\
 & Context & $0.759$ & $61.50$ & $50.15$ & $0.759$ & $70.20$ & $51.11$ & $0.783$ & $68.40$ & $49.90$ & $0.794$ & $62.90$ & $49.69$ & $50.21$ \\
w/o NS & Meaning & $1.000$ & $50.00$ & $50.15$ & $1.000$ & $50.00$ & $50.00$ & $1.000$ & $50.00$ & $49.90$ & $1.000$ & $50.00$ & $49.69$ & $49.94$ \\
 & Context & $0.729$ & $62.20$ & $50.15$ & $0.755$ & $69.60$ & $51.92$ & $0.713$ & $67.80$ & $\underline{51.81}$ & $0.748$ & $66.30$ & $50.00$ & $\underline{50.97}$ \\ \midrule
\multicolumn{2}{c}{XLM-R} & $0.937$ & $59.50$ & $50.10$ & $0.955$ & $69.20$ & $51.41$ & $0.944$ & $67.50$ & $50.00$ & $0.946$ & $62.70$ & $49.69$ & $50.30$ \\
Proposed & Meaning & $0.948$ & $61.10$ & $\underline{60.69}$ & $0.957$ & $66.60$ & $\underline{63.13}$ & $0.950$ & $68.50$ & $\underline{67.04}$ & $0.970$ & $60.70$ & $\underline{51.32}$ & \multicolumn{1}{r}{$\underline{60.54}$} \\
 & Context & $0.916$ & $58.00$ & $49.90$ & $0.941$ & $66.60$ & $50.40$ & $0.936$ & $63.10$ & $49.90$ & $0.933$ & $60.00$ & $49.69$ & \multicolumn{1}{r}{$49.97$} \\
w/o NS & Meaning & $1.000$ & $53.10$ & $\underline{50.30}$ & $1.000$ & $54.90$ & $\underline{51.62}$ & $1.000$ & $54.10$ & $\underline{50.30}$ & $1.000$ & $53.80$ & $\underline{50.10}$ & \multicolumn{1}{r}{$\underline{50.58}$} \\
 & Context & $0.856$ & $59.60$ & $50.10$ & $0.901$ & $69.50$ & $\underline{51.62}$ & $0.877$ & $67.60$ & $\underline{50.10}$ & $0.909$ & $62.70$ & $49.69$ & \multicolumn{1}{r}{$\underline{50.38}$} \\
 \midrule
\multicolumn{2}{c}{mSBERT~\cite{msbert}} & $0.656$ & $69.10$ & $63.00$ & $0.803$ & $70.10$ & $59.49$ & $0.765$ & $76.30$ & $60.58$ & $0.836$ & $67.80$ & $54.68$ & $59.44$ \\
Proposed & Meaning & $0.785$ & $69.80$ & $\bm{64.62}$ & $0.879$ & $69.70$ & $\bm{63.23}$ & $0.772$ & $76.90$ & $\bm{68.25}$ & $0.891$ & $66.20$ & $\bm{63.14}$ & $\bm{64.81}$ \\
 & Context & $0.619$ & $63.10$ & $50.30$ & $0.580$ & $66.20$ & $53.74$ & $0.566$ & $67.30$ & $52.92$ & $0.608$ & $63.40$ & $51.02$ & $52.00$ \\
w/o NS & Meaning & $0.999$ & $57.10$ & $59.07$ & $0.999$ & $59.50$ & $\underline{60.20}$ & $0.999$ & $59.90$ & $56.25$ & $0.999$ & $56.20$ & $\underline{54.99}$ & $57.63$ \\
 & Context & $0.642$ & $68.90$ & $\underline{63.31}$ & $0.795$ & $69.90$ & $59.39$ & $0.752$ & $76.50$ & $\underline{60.69}$ & $0.819$ & $68.00$ & $\underline{55.70}$ & $\underline{59.77}$ \\ \bottomrule
\end{tabular}%
}
\end{table*}

\begin{table}[t!]
\centering
\caption{Results of the STS $2017$ tasks, where ``w/o NS'' denotes our method without negative samples. The best scores are represented in \protect\textbf{bold} fonts, and scores higher than the base pre-trained models are \underline{underlined}.}
\begin{tabular}{@{}ccccccc@{}}
\toprule
 &  & en-ar & en-de & en-tr & es-en & Average \\ \midrule
\multicolumn{2}{c}{mBERT} & $0.178$ & $0.256$ & $0.167$ & $0.197$ & $0.199$ \\
Proposed & Meaning & $0.137$ & $\underline{0.258}$ & $0.000$ & $0.159$ & $0.139$ \\
 & Context & $0.176$ & $0.182$ & $\underline{0.247}$ & $\underline{0.245}$ & $\underline{0.213}$ \\
w/o NS & Meaning & $0.068$ & -$0.014$ & $0.079$ & $0.033$ & $0.041$ \\
 & Context & $\underline{0.197}$ & $\underline{0.273}$ & $\underline{0.191}$ & $\underline{0.211}$ & $\underline{0.218}$ \\ \midrule
\multicolumn{2}{c}{XLM-R} & $0.198$ & $0.215$ & $0.107$ & $0.123$ & $0.161$ \\
Proposed & Meaning & $0.135$ & $0.158$ & $0.017$ & $0.046$ & $0.089$ \\
 & Context & $\underline{0.202}$ & $0.206$ & $\underline{0.170}$ & $\underline{0.173}$ & $\underline{0.188}$ \\
w/o NS & Meaning & $0.008$ & -$0.025$ & $0.055$ & $0.013$ & $0.013$ \\
 & Context & $\underline{0.214}$ & $\underline{0.216}$ & $\underline{0.132}$ & $\underline{0.141}$ & $\underline{0.175}$ \\ \midrule
\multicolumn{2}{c}{mSBERT~\cite{msbert}} & $0.732$ & $0.739$ & $0.722$ & $0.738$ & $0.733$ \\
Proposed & Meaning & $\bm{0.749}$ & $0.731$ & $\bm{0.730}$ & $\bm{0.750}$ & $\bm{0.740}$ \\
 & Context & $0.466$ & $0.540$ & $0.520$ & $0.470$ & $0.499$ \\
w/o NS & Meaning & $0.262$ & $0.312$ & $0.274$ & $0.266$ & $0.278$ \\
 & Context & $\underline{0.735}$ & $\bm{0.740}$ & $\underline{0.729}$ & $\underline{0.744}$ & $\underline{0.737}$ \\ \bottomrule
\end{tabular}%
\label{tab:sts17_results}
\end{table}

\Tref{tab:mcl_wic_results} shows the results of the MCL-WiC tasks: thresholds for making binary judgements, validation set accuracies, and test set accuracies.\footnote{The mBERT model reduced seven test cases on the en-ar task because its tokenizer did not properly support ligatures.} 
Evidently, the language barrier persists in the mBERT and XLM-R models, as shown in the comparable test set accuracies to the majority baseline on the average. 
These findings are consistent with Tiyajamorn~\etal~\cite{nattapoing:emnlp2021} who showed that the language difference separates the spaces of sentence representations of mBERT and XLM-R, and thus representations of semantically similar sentences across languages are far apart. 
The language barrier was alleviated in the mSBERT model that largely outperforms the mBERT and XLM-R. 
Notably, the meaning representations of the proposed method on all the pre-trained models outperformed the mSBERT model averagely. 
Furthermore, the combination of the proposed method and mSBERT achieved the best accuracy, which outperformed mSBERT by $1.61\%$ (en-ar) to $8.45\%$ (en-zh). 
These results imply that the proposed method may be effective in other multilingual pre-trained models, such as LaBSE \cite{labse}. 

\Tref{tab:sts17_results} shows the results of the crosslingual STS tasks. 
Contrary to the monolingual STS tasks, the meaning representations showed inconsistent effects, depending on the pre-trained models. 
The meaning representations improved the mSBERT on three out of four language pairs, whereas they deteriorated the representations of mBERT and XLM-R. 
We conjecture this phenomenon relates to the performance of original pre-trained models; although mSBERT had a $0.733$ Pearson's correlation on average, mBERT and XLM-R had much lower correlations of $0.199$ and $0.161$ to human labels on average, respectively.

\subsection{Ablation Study}
\Tref{tab:result_ls} to \Tref{tab:sts17_results} also show the results of the ablation study, where we excluded the negative samples to train our method. 
Without the negative samples, our method becomes unconstrained; the cross-reconstruction becomes symmetric for the meaning and context distillers. 
Hence, the model loses its ability to distil word meaning in context into the meaning representations. 
This effect was observed for all the tasks in the monolingual and crosslingual settings, where meaning representations without negative samples were no longer useful. 
In the subsequent section, we investigate the characteristics of the meaning and context representations distilled with and without the negative samples.

\section{Analysis}
We investigated the characteristics of the meaning and context representations by analysing the inter-word similarities and similarities to different layers of pre-trained models. 
These investigations further deepened the understanding of the empirical effects of negative samples in our method.

\subsection{Inter-word Similarity Distribution}
We conducted an in-depth analysis using the corpus of paraphrase adversaries from word scrambling (PAWS)~\cite{paws}. 
PAWS is an English paraphrase corpus dedicated to evaluating the sensitivity of recognition models for syntax in paraphrases. 
It provides paraphrase and non-paraphrase pairs that were generated by controlled word swapping and back translation with manual screening. 
Because the pairs in PAWS have relatively high word overlap rates, models that are insensitive to contexts cannot exceed the chance rate for paraphrase recognition.

We generated representations of sentences in the PAWS-Wiki Labelled (Final) section similar to the STS tasks and computed the cosine similarities between them. 
Thereafter, we determined the threshold to regard a pair as a paraphrase using the validation set. 
\Tref{tab:results_paws} shows the results. 
BERT-Large and the transformation method had equal to or lower accuracy than the chance rate of $55.80\%$ (always outputting the majority label of the non-paraphrases). 
In contrast, our method improved the accuracy, even on this challenging task. 
This is achieved by our property that distils word meaning in context preserving useful context information.

\begin{table}[t!]
\centering
\caption{Paraphrase recognition accuracy on challenging PAWS-Wiki corpus}
\label{tab:results_paws}
\begin{tabular}{@{}lcc@{}}
\toprule
           & Threshold & Accuracy (\%) \\ \midrule
All False  & -- & $55.80$ \\
fastText   & $1.000$    & $53.89$         \\
BERT-Large & $0.993$     & $55.76$         \\
Liu~\etal~\cite{liu-etal-2020-towards-better}        & $0.989$     & $55.80$         \\
Proposed       & $0.990$     & $\bm{56.71}$         \\ \bottomrule
\end{tabular}
\end{table}

\begin{table}[t!]
\centering
\caption{Average cosine similarities between words in PAWS-Wiki, where ``w/o NS'' denotes our method without negative samples (``P'' and ``N'' represent paraphrases and non-paraphrases, respectively.) The representations for word meaning in context are expected to have (a) higher similarities for words with the same surfaces than for different words, and (b) higher similarities for words appearing in paraphrases than for words in non-paraphrases.}
\label{tab:results_paws_wordsim}
\begin{tabular}{@{}llcccc@{}}
\toprule
                      &         & \multicolumn{2}{c}{Common words} & \multicolumn{2}{c}{Different words} \\
                      &         & N               & P              & N                & P                \\ \midrule
\multirow{2}{*}{Proposed} & Meaning & $0.712$           & $0.754$          & $0.354$            & $0.374$            \\
                      & Context & $0.806$           & $0.835$          & $0.580$            & $0.595$            \\
\multirow{2}{*}{w/o NS}     & Meaning & $0.998$           & $0.998$          & $0.996$            & $0.996$            \\
                      & Context & $0.705$           & $0.749$          & $0.337$            & $0.357$            \\ \bottomrule
\end{tabular}%
\end{table}

\Tref{tab:results_paws_wordsim} shows the average cosine similarities between representations of common and different words in paraphrases and non-paraphrases of the PAWS-Wiki Labelled (Final) section. 
The representations for word meaning in context are expected to have (a) higher similarities for words with the same surfaces than for different words, and (b) higher similarities for words appearing in paraphrases than for words in non-paraphrases by reflecting the context. 
Particularly, the appropriate representations should have higher similarities for common words in paraphrases than for those in non-paraphrases because the former highly likely has the same meaning.

The meaning and context representations trained with negative samples and the context representations without negative samples preserve these characteristics. 
This indicates that they have a noticeable distinction between common and different words and words in paraphrases and non-paraphrases. 
In contrast, the meaning representations generated without negative samples have high cosine similarities among all the words, regardless of surfaces and paraphrase relations. 
This result implies that the meaning representations without negative samples performed as a noise filter to remove non-useful information from the context representations, and only the corresponding context representations benefited from the training.

The same characteristics can be observed in the crosslingual meaning and context representations. 
Considering \Tref{tab:mcl_wic_results}, the meaning representations obtained without the negative samples required very high thresholds ($1.000$ on mBERT and XLM-R, and $0.999$ on mSBERT) to determine the semantic equivalence.
Together with the lower accuracies on the tasks, these meaning representations approximately lost the ability to distinguish word meanings. 
In contrast, \Tref{tab:sts17_results} shows that the corresponding context representations consistently improve the pre-trained models on the crosslingual STS tasks. 
The improvements confirm that the meaning representations without negative samples perform as a filter, similar to the monolingual settings. 
We conjecture that the filtered information includes language-specific styles.

\subsection{Layer-wise Similarity Distribution}
\begin{figure}[t!]
\centering
\includegraphics[width=0.5\linewidth]{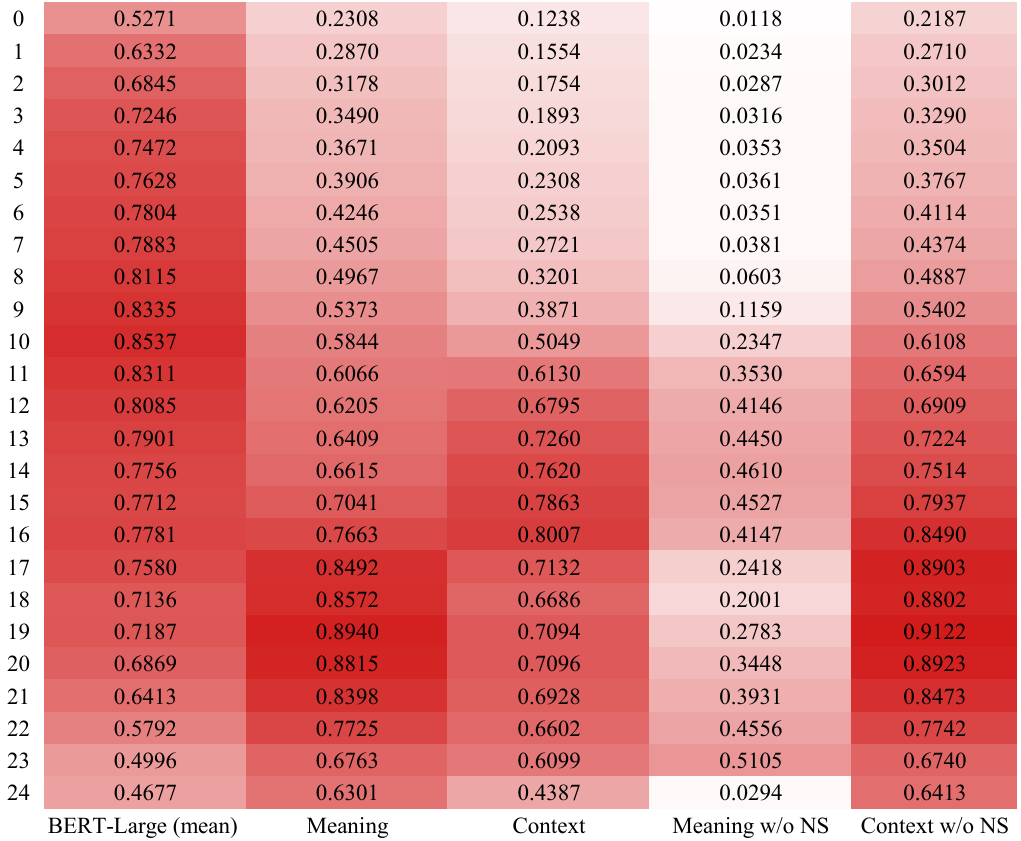}
\caption{Distribution of cosine similarities between monolingual meaning/context representations and BERT-Large layers computed on the SCWS corpus. The darker colours indicate higher similarities.}
\label{fig:layer-wise_sim_monolingual}
\end{figure} 

\begin{figure}[t!]
\centering
\includegraphics[width=0.5\linewidth]{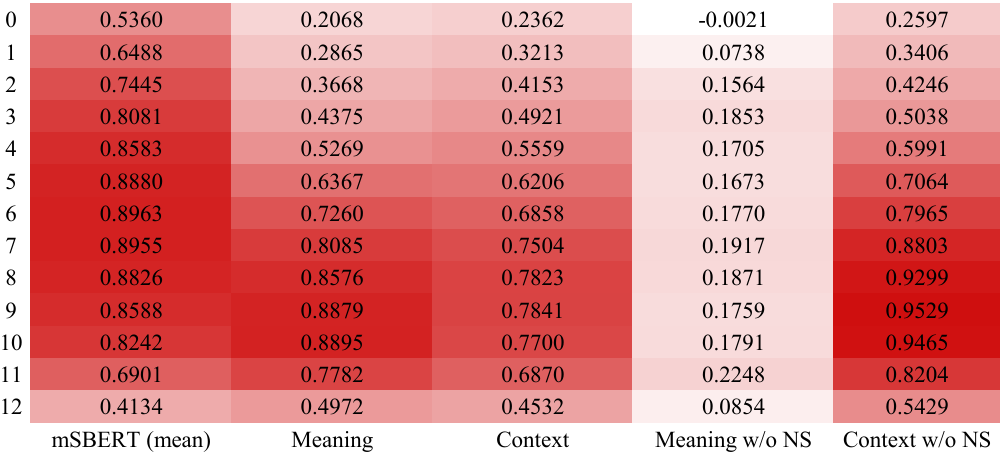}
\caption{Distribution of cosine similarities between crosslingual meaning/context representations and mSBERT layers computed on the MCL-WiC en-ru corpus. The darker colours indicate higher similarities.}
\label{fig:layer-wise_sim_crosslingual}
\end{figure} 

We investigated how our meaning and context representations are similar to the outputs of different layers of pre-trained models. 
\Fref{fig:layer-wise_sim_monolingual} and \Fref{fig:layer-wise_sim_crosslingual} show the average cosine similarities between these representations with or without the negative samples and the corresponding pre-trained models, measured on the SCWS and MCL-WiC en-ru corpora, respectively. 
Specifically, we obtained meaning and context representations of target words and computed the cosine similarities of the outputs of different layers of the pre-trained models of BERT-Large and mSBERT, respectively. 
Considering each figure, the first column shows the mean-pooled top-half layers of the pre-trained model (corresponding to the baselines); the subsequent two columns show the meaning and context representations trained with negative samples; the last two columns show those trained without negative samples. 
Regarding these figures, the darker colours indicate higher cosine similarities. 

The monolingual and crosslingual representations reveal similar characteristics, which are largely different from the mean-pooled original representations.  
The meaning representations trained with negative samples have higher similarities to the upper layers of the pre-trained models, excluding the very top ones. 
In contrast, the context representations are similar to a wider range of layers, middle to upper. 
The lower similarity to the very top layers implies their inferiority as meaning representations, which is consistent with the empirical habits of extracting the second-to-top layers as sentence representations of BERT.\footnote{\url{https://github.com/hanxiao/bert-as-service}}   

In contrast, the meaning representations trained without negative samples have low similarities overall, whereas the layers with relatively higher similarities seem common to the context representations with negative samples. 
This result implies that the meaning representations without negative samples function as a weaker version of the context representations with negative samples. 
These observations confirm that our meaning representations, when trained with negative samples, well combine different layers of the pre-trained model to represent the word meaning in context.

\section{Summary and Future Work}
We have proposed a method that distils a representation of word meaning in context from an off-the-shelf pre-trained masked language model. 
The extensive experimental results showed the effectiveness of the proposed method in both monolingual and crosslingual scenarios. 
Furthermore, the results confirmed that the negative samples are essential as constraints to properly distil the representations of word meaning in context by avoiding degenerated solutions. 

In future work, we plan to investigate the correspondences of the context representations, \ie, what information is distilled in them. 
In addition, we will extend our method to handle low-resource languages in both multilingual and crosslingual settings \cite{liu-etal-2021-am2ico}. 

\section*{Acknowledgment}
This project was funded by the Foundation of Kinoshita Memorial Enterprise.

\bibliographystyle{unsrtnat}
\bibliography{arase_bib}

\appendix
\section{Simple Word Alignment}
\label{appendix:word_alignment}
\Aref{alg:heuristic} shows the word alignment method we used, which is based on cosine similarities between the embedding of words in $S$ and $S_p$. 
Similar simple alignment methods have been commonly used in previous studies \cite{jalili-sabet-etal-2020-simalign,garg-etal-2019-jointly,och-ney-2003}. 
Specifically, we first identify an alignment between words $w_i \in S \setminus w_t$ and $w_j \in S_p$ if and only if they have the highest cosine similarities to each other (\Lref{alg:align}). 
Thereafter, we choose $w_p$ from words that have not been aligned to others and have high cosine similarity to $w_t$ for improved alignment accuracy (\Lref{alg:align_target}) as described below. 

\begin{algorithm} [t!]
\caption{Simple Word Alignment}
\label{alg:heuristic}
\begin{algorithmic}[1] 
\Require Original sentence $S$ containing a target word $w_t$ whose index is $t$, positive sentence $S_p$, static word embedding model $Z$
\Ensure Lexical paraphrase $w_p$ of $w_t$

\State $M \gets \emptyset, A \gets \emptyset, w_p \gets \emptyset$ 
\ForAll{$w_i \in S$ and $w_j \in S_p$} 
    \State $M[i][j] \gets \text{CosineSim}(Z(w_i), Z(w_j))$ \Comment{Compute cosine similarity of embeddings}
\EndFor

\ForAll{$w_i \in S \setminus w_t$} \Comment{Identify alignments of words other than $w_t$}
    \If{$j = \argmax M[i]$ and $i = \argmax M[j]$} \label{alg:align}
        \State $A \gets A \cup \{j\}$
    \EndIf
\EndFor

\ForAll{$j \in \text{argsort}(M[t])$} \Comment{Sort indices in descending order of $M[t]$}
    \If{$j \not\in A$ and \Call{Heuristic}{$M, w_t, w_j$}} \label{alg:align_target} 
        \State $w_p \gets w_j$ 
        \State break;
    \EndIf
\EndFor
\State \Return{$w_p$}

\end{algorithmic} 
\end{algorithm} 

\subsection{Monolingual Case}
An alignment candidate should have a higher cosine similarity to $w_t$ outside the range of $1.0\sigma$ in the distribution of cosine similarities of all the word pairs. 
When $w_p$ is identical or has a similar surface with $w_t$, we also conduct a masked token prediction to obtain more diverse lexical paraphrases. 
Considering the top-$k$ tokens $\hat{T} \in T$ predicted by the masked token prediction whose prediction probabilities are $\hat{Q} \in Q$, where $q_j \in \hat{Q} > \delta$, we replace $w_p$ with $w'_{p} \in \hat{T}$ the word embedding of which has a higher or equal cosine similarity than a pre-determined threshold $\lambda$. 

We also investigated a word substitution approach for creating positive examples~\cite{gari-soler-apidianaki-2020-multisem}, \ie, by replacing only $w_t$ with $w_p$ using masked token prediction. 
This method is computationally faster than the round-trip translation; nevertheless, it shows inferior performance compared to the proposed approach. 
We presume this is because round-trip translation provides more diverse lexical paraphrases compared to those already learned by the masked language model. 
Furthermore, paraphrasing the context enhances the robustness of the meaning and context distillers.

\subsection{Crosslingual Case}
We identify the word alignment $w_p$ of a target $w_t$ from candidates as the one with the highest cosine similarity. 
The candidates should satisfy the conditions: (i) an alignment candidate should have a higher cosine similarity to $w_t$ outside the range of $1.282\sigma$ in the distribution of cosine similarities of all the word pairs, and (ii) the candidate should have a cosine similarity within $50\%$ confidence intervals of cosine similarities of the aligned word pairs $A$.

\section{Filtering of Masked Token Prediction Outputs}
\label{appendix:masked_token_prediction}
We empirically designed heuristics to select $w_n$ while avoiding paraphrases of $w_t$.
\subsection{Monolingual Case}
We used masked token prediction and obtained the top-$k$ predictions of $\hat{T}$ and $\hat{Q}$, where $q_j \in \hat{Q} > \delta$. 
We sorted $\hat{T}$ in a descending order of $\hat{Q}$ and identified $w_n$ the word embedding of which has a lower cosine similarity than $\lambda$. 

We set $\lambda$ as $0.6$ based on the distribution of the cosine similarities of the fastText embeddings on a large text corpus.\footnote{This corpus is independent of this study.} 
We set $k$ as $100$ and $\delta$ as $0.003$ based on the observations of the masked token predictions on several samples randomly extracted from the training corpus, such that we could obtain predictions of reasonable quality. 

\subsection{Crosslingual Case}
We used masked token prediction that masked $w_p$ on $S_p$ with the {\tt [MASK]} label. 
We obtained the top-$k$ tokens $\hat{T} \in T$ as candidates whose prediction probabilities are higher than $\delta$. 
Considering these candidates $w_j \in \hat{T}$, we computed the cosine similarities to $w_t$ and $w_p$, respectively. 
Thereafter, we selected a candidate $w_n$ with the highest masked token prediction probability among the ones that satisfy: (i) a candidate should have a lower cosine similarity to $w_p$ than the average, and (ii) the candidate should have a lower cosine similarity to $w_t$ than $w_p$. 

We set $k$ as $30$ and $\delta$ as $0.001$ based on the observations of the masked token predictions on several samples randomly extracted from the training corpus, such that we could obtain the predictions of reasonable quality.

\section{Dependent Resources}
\label{appendix:resources}
Here is the list of all the URLs of the language resources and libraries used for the study.
\paragraph{Evaluation corpora}
\begin{itemize} 
\setlength{\itemsep}{3pt}
\setlength{\parskip}{0pt}
\item Usim \\ \url{http://www.dianamccarthy.co.uk/downloads/WordMeaningAnno2012/} 
\item WiC \\ \url{https://pilehvar.github.io/wic/} 
\item CoSimlex \\ \url{https://zenodo.org/record/4155986} 
\item SCWS \\ \url{http://www-nlp.stanford.edu/~ehhuang/SCWS.zip} 
\item SentEval \\ \url{https://github.com/facebookresearch/SentEval} 
\item MCL-WiC \\ \url{https://github.com/SapienzaNLP/mcl-wic}
\item STS $2017$ \\ \url{https://public.ukp.informatik.tu-darmstadt.de/reimers/sentence-transformers/datasets/STS2017-extended.zip}
\item PAWS-Wiki Labeled (Final) \\ \url{https://github.com/google-research-datasets/paws} 
\end{itemize}
\paragraph{Language resources}
\begin{itemize}
\setlength{\itemsep}{3pt}
\setlength{\parskip}{0pt}
\item English Wikipedia \\ \url{http://data.statmt.org/wmt20/translation-task/ps-km/wikipedia.en.lid\_filtered.test\_filtered.xz} 
\item WikiMatrix \\ \url{https://github.com/facebookresearch/LASER/tree/main/tasks/WikiMatrix}
\item BERT-large, cased \\ \url{https://huggingface.co/bert-large-cased} 
\item XLM-RoBERTa-large \\ \url{https://huggingface.co/xlm-roberta-large}
\item mBERT-base, cased \\ \url{https://huggingface.co/bert-base-multilingual-cased}
\item XLM-RoBERTa-base \\ \url{https://huggingface.co/xlm-roberta-base}
\item paraphrase-xlm-r-multilingual-v1 \\ \url{https://huggingface.co/sentence-transformers/paraphrase-xlm-r-multilingual-v1}
\item FastText \\ \url{https://dl.fbaipublicfiles.com/fasttext/vectors-english/wiki-news-300d-1M-subword.vec.zip}
\item Aligned fastText \\ \url{https://fasttext.cc/docs/en/aligned-vectors.html}
\end{itemize}
\paragraph{Libraries}
\begin{itemize}
\setlength{\itemsep}{3pt}
\setlength{\parskip}{0pt}
\item WikiExtractor \\ \url{https://github.com/attardi/wikiextractor} 
\item langdetect \\ \url{https://pypi.org/project/langdetect/} 
\item Stanza \\ \url{https://stanfordnlp.github.io/stanza/} 
\item PyTorch (version 1.7.1) \\ \url{https://pytorch.org/} 
\item Transformers (version 4.3.2) \\ \url{https://huggingface.co/transformers/}
\item Implementation of Liu~\etal~\cite{liu-etal-2020-towards-better} \\ \url{https://github.com/qianchu/adjust\_cwe}
\end{itemize}

\end{document}